\documentclass{egpubl}
\usepackage{expressive2019}
 
\Expressive        

\usepackage[T1]{fontenc}
\usepackage{dfadobe}  

\biberVersion
\BibtexOrBiblatex
\usepackage[backend=bibtex,bibstyle=EG,citestyle=alphabetic,backref=true]{biblatex} 
\electronicVersion
\PrintedOrElectronic
\newcommand{\citet}[1]{\cite{#1}}

\ifpdf \usepackage[pdftex]{graphicx} \pdfcompresslevel=9
\else \usepackage[dvips]{graphicx} \fi

\usepackage{egweblnk} 

\usepackage{color,graphicx,amssymb,amsmath,nicefrac}
\usepackage{balance}
\usepackage{subcaption}
\def\mcL{\mathcal{L}}

\def\bbE{{\mathbb E}}
\def\bbR{{\mathbb R}}
\def\lam{\lambda}

\title[Style Transfer]%
      {Learning from Multi-domain Artistic Images  for  Arbitrary Style Transfer}

\author[Zheng Xu et al.]
{\parbox{\textwidth}{\centering Zheng Xu$^{1}$\thanks{contact: xuzh@cs.umd.edu}, Michael Wilber$^{2}$, Chen Fang$^{3}$,  Aaron Hertzmann$^{3}$,   \, and  Hailin Jin$^{3}$
         }
        \\
{\parbox{\textwidth}{\centering $^1$University of Maryland, College Park \, 
           $^2$Cornell Tech, New York \,
			$^3$Adobe  Research, San Jose 
        }
 }
}

\begin{document}


\maketitle
\begin{abstract}
We propose a fast feed-forward network for arbitrary style transfer, which can generate stylized image for previously unseen content and style image pairs. Besides the traditional content and style representation based on deep features and statistics for textures, we use adversarial networks to regularize the generation of stylized images. Our adversarial network learns the intrinsic property of image styles from large-scale multi-domain artistic images. The adversarial training is challenging because both the input and output of our generator are diverse multi-domain images. We use a conditional generator that stylized content by shifting the statistics of deep features, and a conditional discriminator based on the coarse category of styles. Moreover, we propose a mask module to spatially decide the stylization level and stabilize adversarial training by avoiding mode collapse. As a side effect, our trained discriminator can be applied to rank and select representative stylized images. We qualitatively and quantitatively evaluate the proposed method, and compare with recent style transfer methods. We release our code and model at \url{https://github.com/nightldj/behance_release}. 


\ccsdesc[300]{Computing methodologies~ Image manipulation; Non-photorealistic rendering}
\printccsdesc   

\end{abstract}  
\section{Introduction}
Image style transfer is a task that aims to render the content of one image with the style of another, which is important and interesting for both practical and scientific reasons. 
The style transfer techniques can be widely used in image processing applications such as mobile camera filters and artistic image generation.
 Furthermore, the study of style transfer often reveals the intrinsic property of images. 
 Style transfer is challenging as it is difficult to explicitly separate and represent the content and style of an image.  

In the seminal work of \citet{gatys2016image}, the authors represent content with deep features extracted by a pre-trained neural network, and represent style with second order statistics (i.e. the Gram matrix) of the deep features. They propose an optimization framework with the objective that the generated image has similar deep features to the  given content image, and similar second order statistics to the  given style image. The generated  results are visually impressive, but the optimization framework is far too slow for real-time applications.
Later works \cite{johnson2016perceptual,ulyanov2017improved} train a feed-forward network to replace the optimization framework for fast stylization, with a loss similar to \citet{gatys2016image}. However, they need to train a network for each style image and cannot generalize to unseen images. More recent approaches \cite{huang2017arbitrary,li2017universal} tackle arbitrary style transfer for unseen content and style images, which still represent style with second order statistics of deep features. The second order statistics of style representation is originally designed for \emph{textures} \cite{gatys2015texture}, and style transfer is considered as texture transfer in previous methods. 

Another line of research considers style transfer as conditional image generation, and apply adversarial networks to train an image to image translation network \cite{isola2016image,taigman2016unsupervised,zhu2017unpaired,huang2018multimodal}. The trained image translation networks can transfer image from one domain to another domain, for example, from a natural image to sketch. However, they cannot be applied to arbitrary style transfer as the input images are from mutliple domains.



In this paper, we combine the best of both worlds by adversarially training a single feed-forward network for arbitrary style transfer. 
We introduce several techniques to tackle the challenging problem of adversarial training from multi-domain data. 
In adversarial training, the generator (stylization network) and the discriminator are alternatively updated. 
Both our generator and discriminator are conditional networks. 
The generator is trained to fool the discriminator, as well as satisfy the content and style representation similarity to inputs.
 Our generator is built upon a state-of-the-art network for abitrary style transfer \cite{huang2017arbitrary}, which is conditioned on both content image and style image, and uses adaptive instance normalization (AdaIN) to combine the two inputs. AdaIN shifts the mean and variance of the deep features of content image to match those of the style image.  
 Our discriminator is conditioned on the coarse domain categories, which is trained to distinguish the generated images with real images from the the same style category. 
 
 Comparing with previous arbitrary style transfer methods, our approach uses the discriminator to learn a data-driven representation for styles.
 The combined loss for our generator considers both instance-level information from style loss and category-level information from adversarial training. 
   Comparing with previous adversarial training methods, our approach handles multi-domain inputs by using a conditional generator designed for arbitrary style transfer and a conditional discriminator.  
Moreover, we propose a mask module to automatically control the level of stylization by predicting a mask to blend the stylized features and the content features. 
Finally, we use the trained discriminator to rank and find the representative generated images in each style category. We release our code and model at \url{https://github.com/nightldj/behance_release}

\section{Related work}

\textbf{Style transfer. }
We briefly review the neural style transfer methods, and recommend \cite{jing2017neural} for a more comprehensive review.  
\citet{gatys2016image} proposed the first neural style transfer method based on an optimization framework, which uses deep features to represent content and Gram matrix to represent style. 
The optimization framework was replaced by a feed forward network to achieve real-time performance in \cite{johnson2016perceptual,ulyanov2016texture,wang2017multimodal}. \citet{ulyanov2017improved} showed that instance normalization is particularly effective for training a fast style transfer network.  
Other works focused on controlling spatial, color, and stroke for  stylization \cite{gatys2017controlling,frigo2016split,jing2018stroke}, and exploring other style representation such as mean and variance \cite{li2017demystifying}, histogram \cite{wilmot2017stable}, patch-based MRF \cite{li2016combining}, and patch-based GAN \cite{li2016precomputed}. 
Comparing with \cite{gatys2016image},  these fast style transfer methods sometimes compromise on the visual quality, and need to train one network for each style. 

Various methods have been proposed to train a single feed forward network for multiple styles. 
\citet{dumoulin2016learned} proposed conditional instance normalization, which learned the affine parameter for each style image.  
\citet{chen2017stylebank} learned the ``style bank'', which contains several layers of filters for each style.  
\citet{zhang2017multi} proposed comatch layers for multi-style transfer. These methods only work with limited number of styles, and cannot apply to an unseen style image.

More recent approaches are designed for arbitrary style transfer, where both the content and the style  inputs can be unseen images. 
\citet{ghiasi2017exploring} extended conditional instance normalization (IN) by training a separate network to predict the affine parameter of IN.
\citet{shen2018style} learned a meta network to predict filters in the transformation networks. \citet{huang2017arbitrary} proposed adaptive instance normalization (AdaIN) that adjusts the mean and variance of content image to match those of the style image. \citet{li2017universal,li2018closed} used feature whitening and coloring transforms (WCT) to match the statistics of the content image to the style image. \citet{sheng2018avatar} proposed feature decoration that generalizes AdaIN and WCT. Note that the optimization framework \cite{gatys2016image} and  path-based non-parametric methods (e.g., style swamp \cite{chen2016fast}, deep image analogy\cite{liao2017visual}, and deep feature reshuffle \cite{gu2018arbitrary}) can also be applied to arbitrary style transfer, but these methods can be much slower.
\citet{zhang2017separating} proposed to separate style and content and then combine them with bilinear layer, which requires a set of content and style images as input and has limited applications.
Our approach is the first to explore adversarial training for arbitrary style transfer.

\textbf{Generative adversarial networks (GANs).} GANs have been widely studied for image generation and manipulation tasks since \cite{goodfellow2014generative}. 
\citet{elgammal2017can} applied GANs to generate artistic images.  
\citet{isola2016image} used conditional adversarial networks to learn the loss for image to image translation, which is extended by several concurrent methods \cite{zhu2017unpaired,kim2017learning,yi2017dualgan,liu2017unsupervised} that explored cycle-consistent loss when  training data are unpaired. Later works improved the diversity of generated images by considering multimodality of data \cite{zhu2017toward,almahairi2018augmented,huang2018multimodal}. Similar techniques have been applied to specific image to image translation tasks such as image dehazing \cite{yang2018towards}, 
face to cartoon \cite{taigman2016unsupervised,royer2017xgan} and font style transfer \cite{azadi2017multi}. These methods successfully train a transformation network from one image domain to another. However, they cannot handle multi-domain input and output images, and it is known to be difficult to generate images with large variance \cite{chen2016infogan,odena2016conditional,miyato2018cgans}. Our approach adopt conditional generator and discriminator to tackle  the multi-domain input and output for arbitrary style transfer. 



\begin{figure*}[t]
\centerline{
\includegraphics[width=1.01\linewidth]{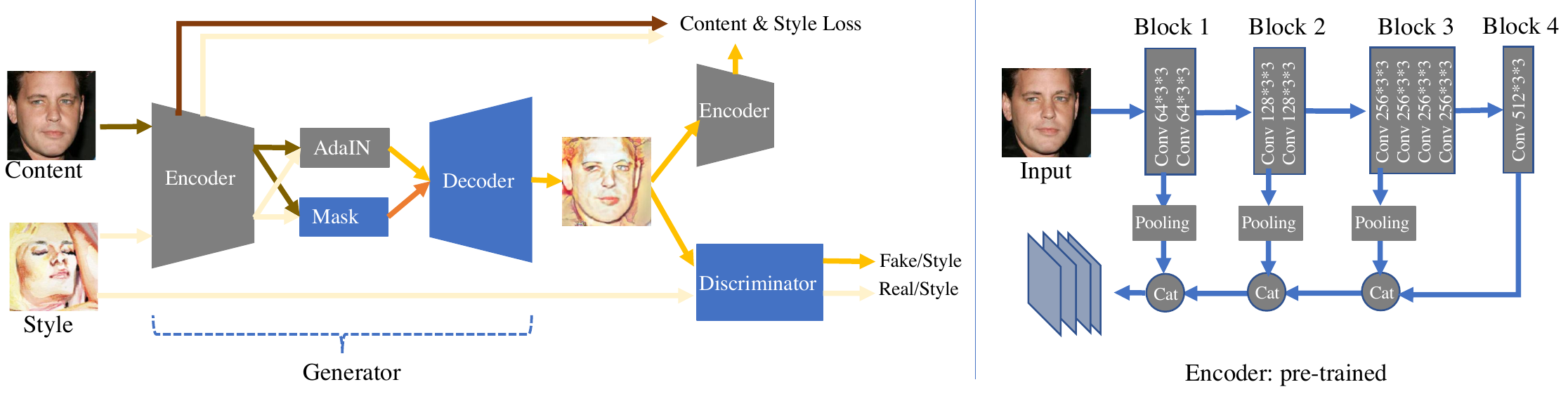}
}
\caption{
Proposed network: (left) encoder-decoder as generator; (right)  pre-trained VGG as encoder. The decoder architecture is symmetric comparing to encoder. We use the conventional texture loss based on pre-trained encoder features, and adversarially train mask module, decoder and discriminator.}
\label{fig:net}
\end{figure*}

\section{Proposed method}
We use an encoder-decoder architecture as our transformation network, and use the convolutional layers of the pre-trained VGG net \cite{simonyan2014very,xu2018effectiveness} as our encoder to extract the deep features. We add skip connections and concatenate the features from different levels of convolutional layers as the output feature of the encoder. We adopt adaptive instance normalization (AdaIN) \cite{huang2017arbitrary} to adjust the first and second order statistics of the deep features. 
Furthermore, we generate spatial masks to automatically adjust the stylization level. 
Our transformation network is a conditional generator inspired by the state-of-the-art network for arbitrary style transfer. 
Our network is trained with perceptual loss for content representation, Gram loss for style representation as in \cite{gatys2016image,johnson2016perceptual,ulyanov2016texture}, as well as the adversarial loss to capture the common style information beyond textures from a style category. We show the proposed network in figure \ref{fig:net}, and provide details in the following sections. 

\subsection{Network architecture} 
Our \textbf{encoder} uses the convolutional layers of the VGG net \cite{simonyan2014very} pre-trained on Imagenet large-scale image classification task \cite{russakovsky2015imagenet}. 
VGG net contains five blocks of convolutional layers, and we adopt the first three blocks and the first convolutional layer of the forth block. 
Each block contains convolutional layers with ReLU activation \cite{krizhevsky2012imagenet}, and 
the width (number of channels) and size (height and width) of the convolutional layers are shown in figure \ref{fig:net}. 
There is a maxpooling layer of stride two between blocks, and the width of convolutional layer is doubled after the downsampling by maxpooling.
We concatenate the features from the first convolutional layer of each block as the output of the encoder. These skip connections help to transfer style captured by both high-level and low-level features, as well as make the training easier by smoothing the loss surface of neural networks \cite{li2018visualizing}. 

Our \textbf{decoder} is designed to be almost symmetric to the VGG encoder, which has four blocks and between blocks are transposed convolutional layer for upsampling. We add LeakyReLU \cite{he2015delving} and batch normalization \cite{ioffe2015batch} to each convolutional layer for effective adversarial training \cite{radford2015unsupervised}. The decoder is trained from scratch. 

\textbf{Adaptive instance normalization (AdaIN)} has been shown to be effective for image style transfer~\cite{huang2017arbitrary}. AdaIN shifts the mean and variance of deep features of content to match style with no learnable parameters. 
Let $x, y\in \bbR^{N\times C\times H\times W}$ represent the features of a convolutional layer from a minibatch of content and style images, where $N$ is the batch size, $C$ is the width of the layer (number of channels), $H$ and $W$ are height and width, respectively. $x_{nchw}$ denotes the element at height $h$, width $w$ of the $c$th channel from the $n$th sample, and adaIN layer can be written as,
\begin{equation}
A_{nchw}(x, y)  = \sigma_{nc}(y) \left( \frac{x_{nchw} - \mu_{nc}(x)}{\sigma_{nc}(x)} \right) + \mu_{nc}(y)
\end{equation}
where $ \mu_{nc} (x)  = \nicefrac{1}{HW} \sum_{h,w=1}^{H,W}  x_{nchw}$,  $\sigma_{nc} (x)= \sqrt{\nicefrac{1}{HW} \sum_{h,w=1}^{H,W}  (x_{nchw}-\mu_{nc})^2 + \epsilon}$, $\epsilon$ is a very small constant, and $\mu_{nc}(x), \sigma_{nc}^2(x)$ represent the mean and variance for the $c$th channel of the $n$th sample of feature $x$. 

The \textbf{mask} module in our network contains a few convolutional layers operated on the concatenation of content feature $x$ and style feature $y$. The output is a spatial soft mask $M(x, y) \in [-1, 1]^{N\times C\times H\times W}$ that has the same size as feature and each value is between $-1$ and $1$. The generated mask $M(x, y)$ is used to control the stylization level by linearly combine the adaIN feature $A(x, y)$ and the original content feature $s$ as the input of the decoder,
\begin{equation}
z = M(x,y) \times x + (1 - M(x,y)) \times A(x,y),
\end{equation}
where the element-wise operations are used for combining these features. 

Our \textbf{discriminator} is a patch-based network inspired by \cite{isola2016image}. To handle the multi-domain images for arbitrary style transfer, our discriminator is conditioned on the style category labels. Inspired by AC-GAN \cite{odena2016conditional}, our discriminator predicts the style category and distinguish the real image and fake image at the same time. We also adopt the projection discriminator  \cite{miyato2018cgans} to make sure the style category conditioning will not be ignored.

\subsection{Adversarial training}
We alternatively update the generator (mask module and decoder) and discriminator during training, and apply prediction optimizer \cite{yadav2017stabilizing} to stabilize the training.

\textbf{ Generator update.} Our generator takes a content image and a style image as input, and outputs the stylized image. 
The generator is updated by minimizing the loss combined of adversarial loss $\mcL_A$, style classification loss $\mcL_{DS}$, content loss $\mcL_c$ and style loss $\mcL_s$,
\begin{equation}%
\min_{G} \mcL_{G} = \mcL_A + \lam_{DS} \mcL_{DS} + \lam_{c} \mcL_c + \lam_{s} \mcL_s, \label{eq:combine}
\end{equation}%
where $\lam_{DS}, \lam_{c}, \lam_{s}$ are hyperparameters for the weights of different losses. Let us denote the feature map of the $l$th layer in our encoder as $x^{(l)}, y^{(l)}$,  the input content and style images as $x^{(0)}, y^{(0)}$,  the generator network as $G(\cdot,\cdot)$, and the discriminator network as $D(\cdot)$. 
When the discriminator $D(\cdot)$ is fixed, the output stylized images $\hat x = G(x^{(0)}, y^{(0)})$ aim to fool the discriminator, and also be classified to same style category $s$ as the input style image,
\begin{equation}%
\begin{split}
 \mcL_A & = \bbE[\log \text{Prob}(\text{Real} | D(\hat x))], \\
  \mcL_{DS} & = \bbE[\log \text{Prob}(s | D(\hat x))]. 
  \end{split}
\end{equation}%
 $\mcL_A$ and $\mcL_{DS}$ are learned loss that capture the category-level style of images from the training data. We also use the traditional content and style loss based on deep features and Gram matrix,
 \begin{equation}%
 \begin{split}
 \mcL_c & = \bbE[\| x^{(4)} - \hat x^{(4)}\|_1], \\
   \mcL_{s} & = \bbE[\sum_{l=1}^{4} \| \text{Gram}(y^{(l)}) - \text{Gram}(\hat x^{(l)}) \|_1]. 
    \end{split}
\end{equation}%
We use the deep feature from the forth block of pre-trained VGG net for content representation, and use the Gram matrix from all the blocks for style representation.  We find $\ell_1$ norm is more stable than $\ell_2$ when combining with the adversarial loss.

\begin{figure*}[t]%
\centerline{
\includegraphics[width=0.7\linewidth]{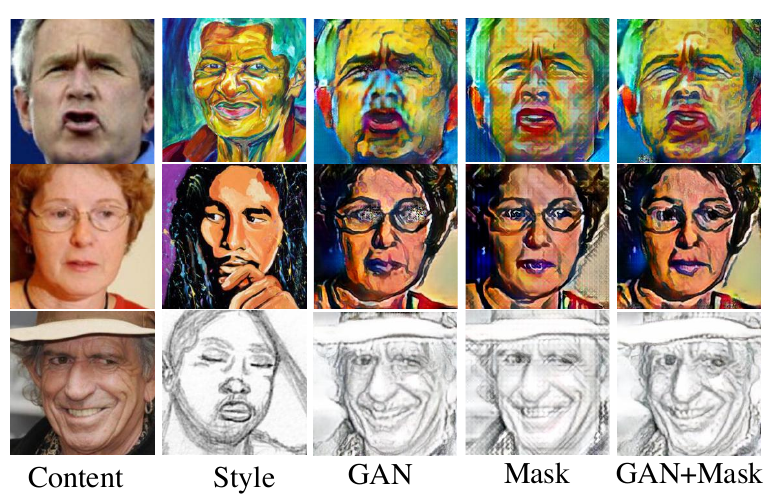}
}
\caption{Benefits of adversarial training and mask module. We show the encoder-decoder network with adversarial training only,  mask module only, and the combination of adversarial training and mask module. Mask module only does not improve the visual quality of generated images, which have artifacts and undesired textures. GAN only can generate collapsed images with corrupted eyes and noses. }
\label{fig:abl}
\end{figure*}%

\textbf{Discriminator update.} Our discrimintor is conditioned on style category to handle the multi-domain generated images, inspired by \cite{chen2016infogan,odena2016conditional,miyato2018cgans,xu2018training}. When the generator is fixed, the discriminator is adversarially trained to distinguish the generated images and the real style images,
\begin{equation}
\min_{D} \mcL_{D} = \hat \mcL_A +  \lam_{DS} \hat \mcL_{DS}, 
\end{equation}  
where {\small  $ \hat  \mcL_A = \bbE[\log \text{Prob}(\text{Fake} | D(\hat x)) + \log \text{Prob}(\text{Real} | D(y))]$},  and {\small   $\hat \mcL_{DS} = \bbE[\log \text{Prob}(s | D(\hat x)) +\log \text{Prob}(s | D(y))]$ }. 

\textbf{Discriminator for ranking.} The adversarilly trained discriminator characterizes the real style images, and hence can be used to rank the generated images. We rank the stylized images $\hat x$ based on the likelihood score $\text{Prob}(s | D(\hat x)) *\text{Prob}(\text{Real} | D(\hat x))$.

\subsection{Ablation study}
The encoder-decoder architecture and adaIN module have been shown to be effective in previous work \cite{huang2017arbitrary}.  We use visual examples to show the importance of mask module and adversarial training in the proposed method in figure \ref{fig:abl}. 
We present results from adversarially trained network without mask module, network with mask module but trained without adversarial loss, and the proposed method.
When trained without adversarial loss,  the network produces visually similar results with or without mask module as the network is over-parameterized. 

 \begin{figure*}[h]
\centerline{
\includegraphics[width=0.87\linewidth]{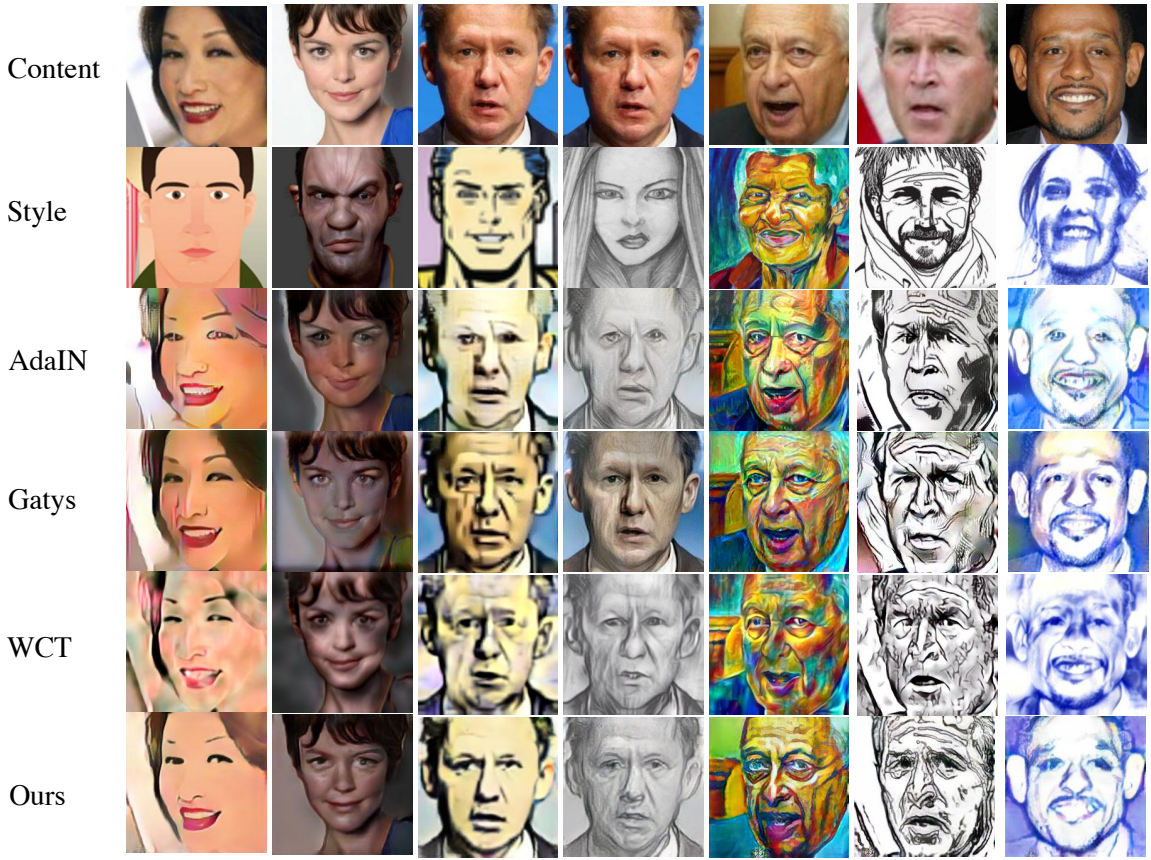}
}
\caption{
Qualitative evaluation for style transfer. We shown examples of transferring photos to seven different styles. AdaIN and WCT will generate artifacts and undesired textures. Gatys' results are more visually appealing, but the optimization is slow, and it is hard to choose the parameter to control stylization level. Our method efficiently generate clean and stylized images.}
\label{fig:vis}
\end{figure*}%

\begin{table*}[tbhp]
\centering
\caption{Quantitative evaluation for style transfer. Our method is preferred by human annotators and outperforms baselines.} 
\setlength\tabcolsep{4pt}
\begin{tabular}{|c|c|c|c|c|c|c|c|c|c|}
\hline
 & vectorart & 3D graphics & comic & graphite & oil paint & pen ink & water color & all \\
\hline
AdaIN~\cite{huang2017arbitrary} & 0.2849 &   0.2029  &  0.2314  &  0.1277  &  0.3018 &   0.2151  &  0.2118  &  0.2199 \\
\hline
WCT~\cite{li2017universal} &      0.1134 &   0.1957 &    0.2066 &   \textbf{0.4754}  &  0.3350  &  0.2868  &  \textbf{0.4409}  &  0.3001 \\
\hline
Ours &  \textbf{0.6017}  &  \textbf{0.6014}  &  \textbf{0.5620} &   0.3969 &   \textbf{0.3632}  &  \textbf{0.4981}  &  0.3473  &  \textbf{0.4800} \\
\hline
\end{tabular}%
\label{tab:style}%
\end{table*}%

Our adversarial training significantly improves the visual quality of the generated images in general. The block effects
 and many other artifacts are removed through adversarial training, which makes the generated images look more ``natural''. 
 Moreover, the data-driven discriminator learns to distinguish foreground and background well; adversarial training cleans the background and adds more details to the foreground. 
 Our mask module controls the stylization level at different spatial location of the image, which significantly improves the stylization of salient components like eyes, nose and mouth of a face. 
 The salient regions are repeatedly captured by the deep features from high-level layers, which can make them difficult to handle when adjusting the statistics of the features.  
 By controlling the stylization level, the mask module prevents over-stylization of salient region, and also helps adversarial training by relieving the mode collapse of salient regions.

\section{Experiments} \label{sec:exp}
We qualitatively and quantitatively evaluate the proposed method with experiments. We extensively use the Behance dataset~\cite{wilber2017bam} for training and testing. Behance~\cite{wilber2017bam} is a large-scale dataset of artistic images, which contains coarse category labels for content and style. We use the seven media labels in Behance as style category:  vector art, 3D graphics, comic, graphite , oil paint, pen ink, and water color. We create four subsets from the Behance images for face, bird, car, and building. Our face dataset is created by running a face detector on a subset of images with people as content label and contains roughly 15,000 images for each style. The other three are created by selecting the top 5000 ranked images of each media for the content, respectively. We add describable textures Dataset (DTD)~\cite{cimpoi14describing} as another style category to improve the robustness of our method. We add natural images as both content images and an extra style for each subset. Specifically, we use labeled faces in the wild (LFW)~\cite{LFWTech}, the first 16,000 images of CelebA dataset~\cite{liu2015faceattributes}, Caltech-UCSD birds dataset~\cite{WelinderEtal2010}, cars dataset~\cite{KrauseStarkDengFei-Fei_3DRR2013}, and Oxford building dataset~\cite{Philbin07}. In total, we have nine style categories in our data. We split both content and style images into training/testing set, and use unseen testing images for our evaluation. The total number of training/testing images are 122,247 / 11,325 for face,  35,000 / 3,505 for bird, 36,940 / 3,700 for car, and 34,374 / 3,444 for building. 

We train the network on face images, and then fine-tune it on bird, car, and building. We use Adam optimizer with prediction method~\cite{yadav2017stabilizing} with learning rate $2e-4$ and parameter $\beta_1=0.5, \beta_2=0.9$. We train the network with batch size 56 for 150 epochs and linearly decrease the learning rate after 60 epochs. It takes about 8 hours to complete on a workstation with 4 GPUs. We set all weights in our combined loss \eqref{eq:combine} as $1$ except for $\lam_{s} = 200$ for the style loss. The weights are chosen so that different components of the loss have similar numerical scales. The training code and pre-trained model in Pytorch are released in \url{https://github.com/nightldj/behance_release}. 

We compare with arbitrary style transfer methods, the optimization framework of neural style transfer (Gatys)~\cite{gatys2016image}, and two state-of-the-art methods, adaptive instance normalization (AdaIN)~\cite{huang2017arbitrary} and feature transformation (WCT)~\cite{li2017universal}. Note that our approach, AdaIN and WCT apply feed-forward network for style transfer, which are much faster than Gatys method.

\begin{figure*}[h]%
\centerline{
\includegraphics[width=0.9\linewidth]{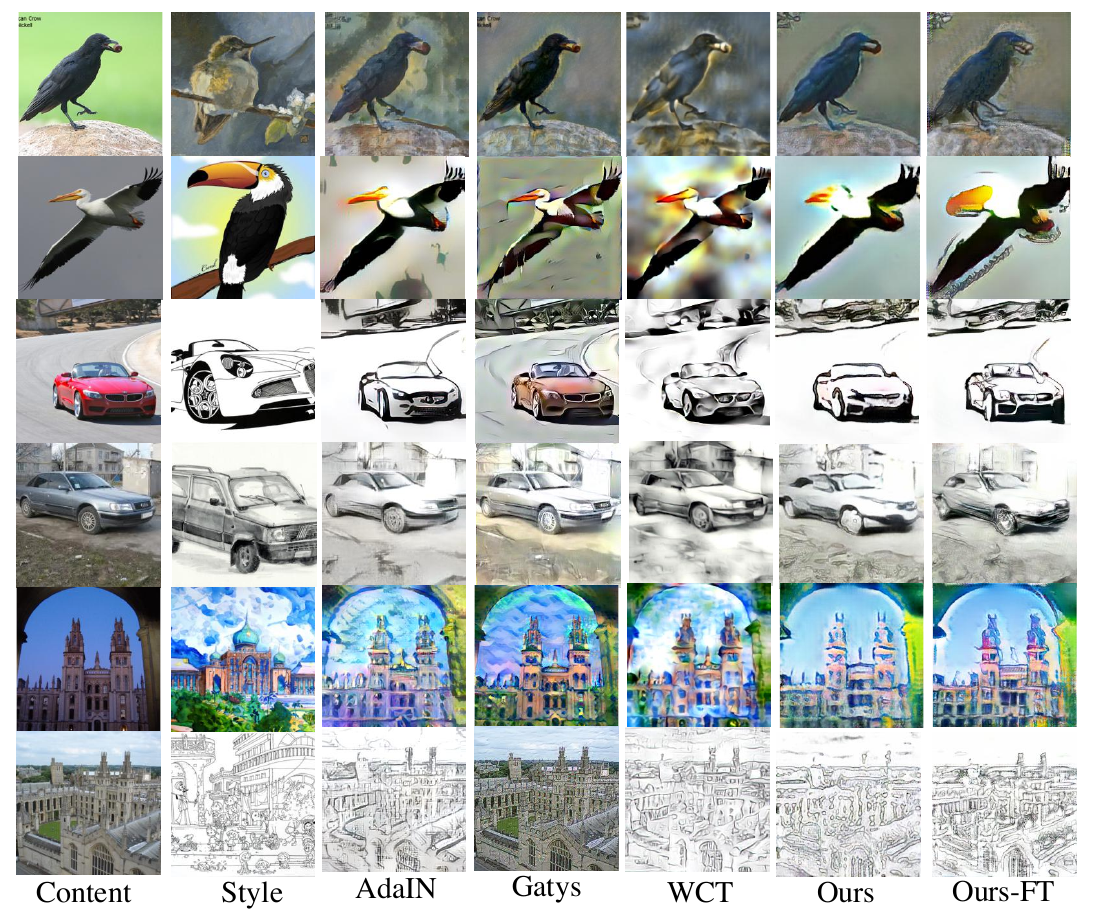}
}
\caption{
Qualitative evaluation for general objects. This task is more difficult for our GAN-based method because the training data is more noisy, especially for bird images with large diversity. Our method can generate clean background, detailed foreground, and better stylized strokes. }
\label{fig:obj}
\end{figure*}%

\begin{table*}[tbhp]
\centering
\caption{Quantitative evaluation for style transfer of building. Different methods are competitive for different styles. The overall performance of our method is better.} 
\setlength\tabcolsep{4pt}
\begin{tabular}{|c|c|c|c|c|c|c|c|c|c|}
\hline
 & vectorart & 3D graphics & comic & graphite & oil paint & pen ink & water color & all \\ 
\hline
AdaIN~\cite{huang2017arbitrary} & 0.2119 &   0.2703  &  0.3089 &    0.3260 &   0.2778 &   \textbf{0.3944} &   0.3654  &  0.3203  \\
\hline
WCT~\cite{li2017universal} &   \textbf{0.4503}  &  \textbf{0.4865} &   \textbf{0.3740} &   0.1547  &  \textbf{0.4383}  &  0.2310 &   0.1731  &  0.3145 \\
\hline
Ours &      0.3377 &   0.2432  &  0.3171 &   \textbf{0.5193}  &  0.2840 &    0.3746 &   \textbf{0.4615} &  \textbf{0.3652} \\
\hline
\end{tabular}%
\label{tab:obj}%
\end{table*}%

\begin{figure*}[t]%
\centerline{
\includegraphics[width=0.8\linewidth]{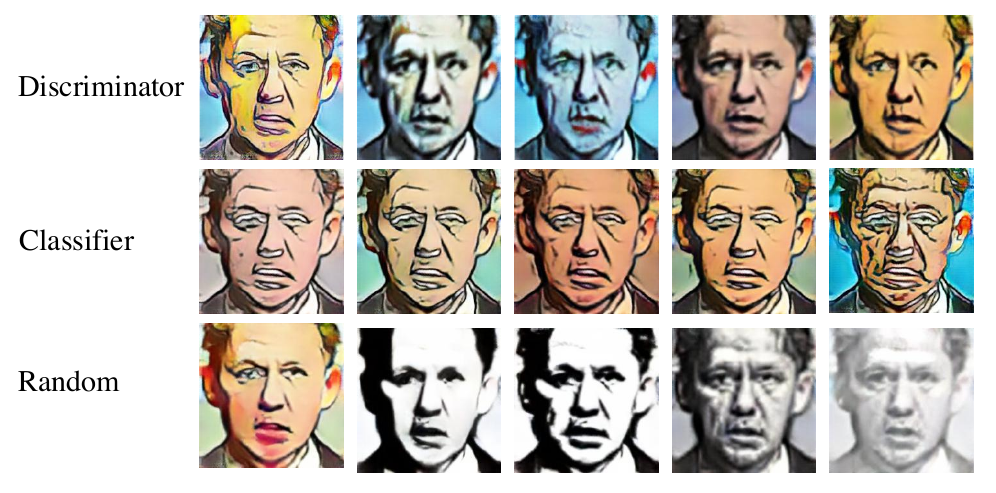}
}
\caption{
Qualitative evaluation for style ranking.}
\label{fig:rank}
\end{figure*}%

\subsection{Evaluation of style transfer}
We qualitatively compare our approach with previous arbitrary style transfer methods, and present some results in figure \ref{fig:vis}. We show seven pairs of content and style images from our face dataset, and the style images are from testing set of vector art, 3D graphics, comic, graphite , oil paint, pen ink, and water color, respectively. For Gatys method~\cite{gatys2016image}, we tune the weight parameter, and select the best visual results from either Adam or BFGS as optimizer. For AdaIN~\cite{huang2017arbitrary} and WCT~\cite{li2017universal}, we use their released best models. The content and style images are from the separate testing set that have not been seen for our approach and the baseline methods.

Gatys method~\cite{gatys2016image} is sensitive to parameter and optimizer setting. We may get results that are not stylized enough even after parameter tuning due to the difficulty of optimization. AdaIN~\cite{huang2017arbitrary} often over-stylizes the content image, creates undesirable artifacts, and sometimes changes the semantic of the content image. WCT~\cite{li2017universal} suffers from severe block effect and artifacts. The previous methods all create texture-like artifacts because of the texture-based style representation. For example, the stylized images of baselines in the first column of figure \ref{fig:vis} have stride artifacts. Our approach generate more visually appealing results with clean background, vivid foreground, and more consistent with the style of the input.

We conduct user study on Amazon Mechanical Turk, and present quantitative results in table \ref{tab:style}. We compare with the two recent fast style transfer methods in this study. We randomly select 10 content images and 10 style images from each Behance style category to generate 700 testing pairs. For each pair, we show the stylized images by our approach, AdaIN~\cite{huang2017arbitrary}, and WCT~\cite{li2017universal}, and ask 10 users to select the best results. We remove the unreliable results that are labeled too soon, and show preference (click) ratio for different style categories. WCT~\cite{li2017universal} performs well on graphite and water color, where the style images themselves are visually not ``clean''. Our approach achieves the best results in the other five categories and is overall the most favorable.


\begin{figure}[tbhp]%
\centerline{
\includegraphics[width=0.95\linewidth]{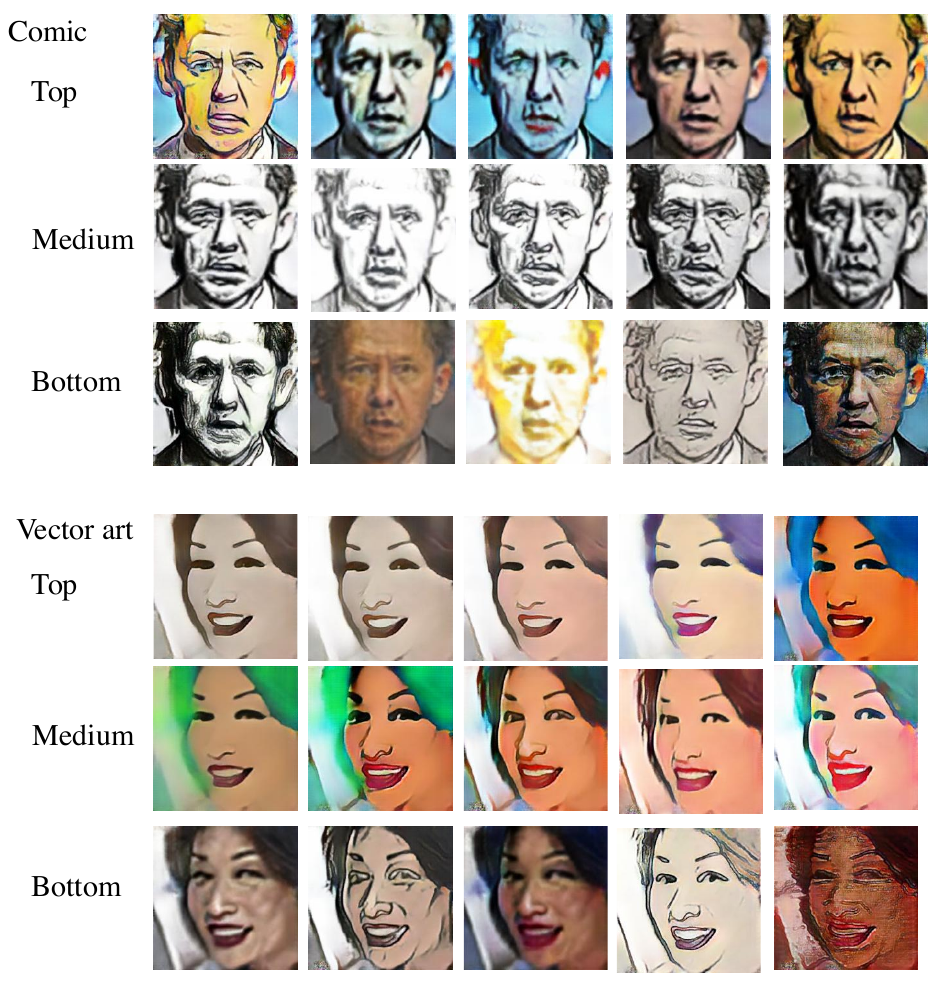}
}
\caption{
Ranking stylized images by our discriminator.}
\label{fig:r2}
\end{figure}%

\subsection{Evaluation of style transfer for general objects}
We evaluate the performance of the proposed approach on general objects beyond face. Specifically, we test for bird, car, and building. In figure \ref{fig:obj}, we show the stylized images generated by our network trained on face (Ours), as well as fine-tuned for each object (Outs-FT). 
Our network trained on face generalizes well, and generates images  look comparable, if not better than, the baseline methods. 
Fine-tuning on bird does not help the performance. The adversarial training may be too difficult for bird because the given training style images are noisy and diverse. 
Fine-tuning on car and building brings more details to the foreground object of our generated images.
The training images of car and building are also noisy and diverse, but these objects are more structured than bird. We show more results on our performance on general object tasks in the supplementary material.

We conduct the user study for building images and report results in table \ref{tab:obj}. Our approach achieves good results for graphite and water color because of the clean background in our generated images. For the other categories, our results are comparable with baselines. Our overall performance is still the best.

\subsection{Evaluation for style ranking}

\begin{figure*}[t]%
\centerline{
\includegraphics[width=0.8\linewidth]{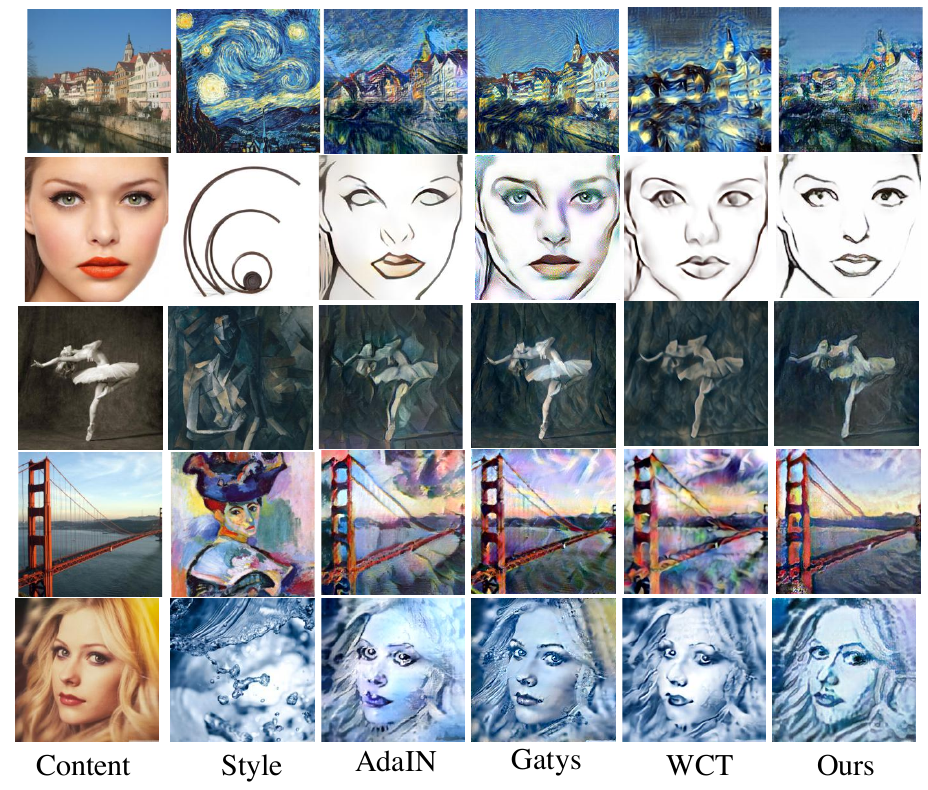}
}
\caption{
Qualitative evaluation for style transfer on texture-centric cases in previous papers. Our method generates stylized images with clean background, which are visually competitive to the previous methods that targeted only on texture transfer.}
\label{fig:gen}
\end{figure*}%

We apply the trained discriminator to rank the generated images for a style category. 
Figure \ref{fig:rank} show the top five generated images by stylizing with all the testing images in comic style. The stylized images are generated by our network, and ranked by our discriminator, a style classifier, and random selection, respectively.  The style classifier use the same network architecture as our discriminator  and training data as our method. The hyper parameters are tuned to achieve the best style classification accuracy on the separate validation dataset, which makes the style classifier a strong baseline. Our generator network produced good results, and even random selected images look acceptable. The top selected results of our discriminator are more diverse, and more consistent to the comic style because of the adversarial training. 

Figure \ref{fig:r2} shows more ranked images  by our discriminator at top, in the middle, and at the bottom for two content images stylized by images from two categories.  The top ranked results are more visually appealing, and more consistent with the style category.

Finally, we conduct user study to compare the ranking performance of our discriminator and the baseline classifier. We generated images by stylizing ten content images with all the testing images for each of the seven Behance styles, and rank the 70 sets of results. We comparing the rank of each generated image by discriminator and classifier, and select five images that are ranked higher by our discriminator, and five images that are ranked higher by the baseline classifier. We show the ten images to ten users and ask them to select five images for each set. The preference ratio of our discriminator is $0.5068$ comparing to $0.4932$ of classifier. We beat a strong baseline in a highly subjective and challenging evaluation.


\section{Supplemental experiments}
In this section, we present supplemental experiments to show some side effect of the proposed method. We first demonstrate our method can be applied to previous style transfer test cases which focus on transferring textures of the style image. We then show that the proposed method can be applied to destylization and generate images look more realistic than baselines.

\subsection{Examples for general style transfer}
In figure \ref{fig:gen}, we evaluate on test cases from previous style transfer papers. The style images have rich texture information, and the content images vary from face to building. Our network is trained on our face dataset described in section \ref{sec:exp}. Our network generalizes well and produces comparable results, if not better than, comparing with baselines. Particularly, our approach often generates clean background without undesired artifacts.

\begin{figure*}[t]%
\centerline{
\includegraphics[width=0.75\linewidth]{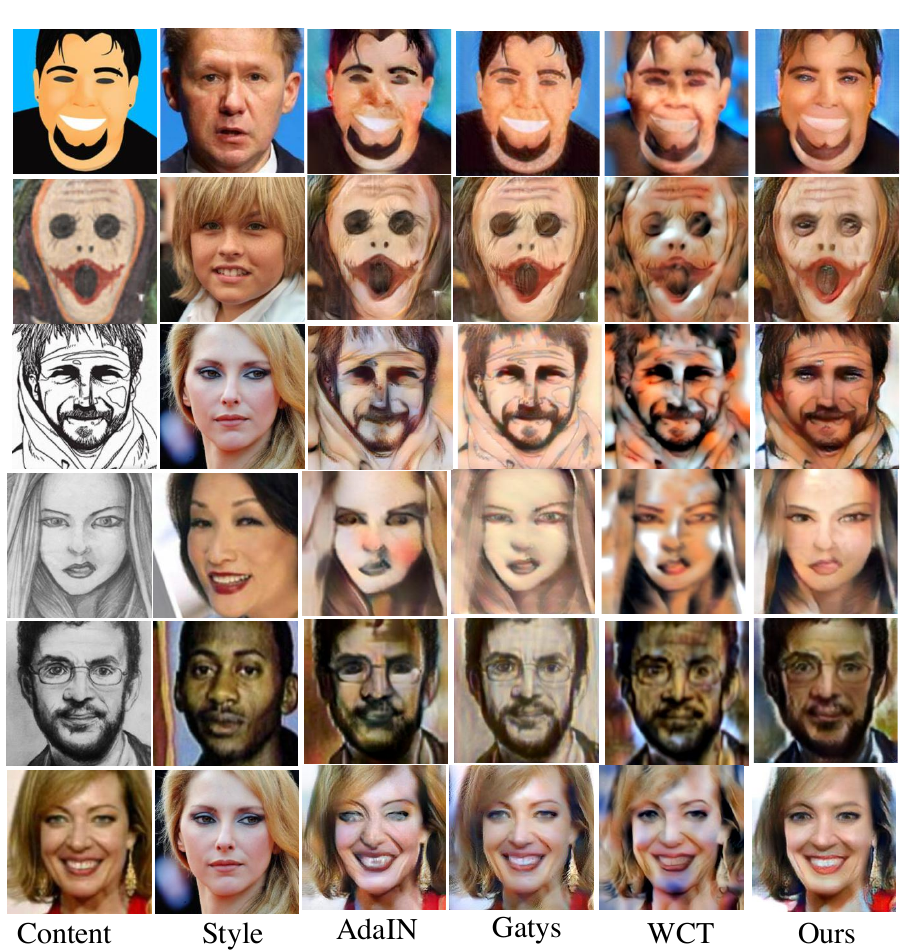}
}
\caption{
Qualitative evaluation for destylization.}
\label{fig:des}
\end{figure*}%
\subsection{Destylization}
We show that if we also use artistic images as content images during training, the exact same architecture can be used to destylize images (figure \ref{fig:des}).
Destylization is a difficult task because we only use one network to destylize diverse artistic images.
The training also becomes much more difficult as the number of pairs increase square to the samples. 
Though there is still room to improve, our adversarial training and network architecture look promising in limited training time. 
The last row in \ref{fig:des} also suggests our network can transfer style of photorealistic images, which is difficult for the baselines.  

\section{Conclusion and discussion}
We propose a feed-forward network that uses adversarial training to enhance the performance of arbitrary style transfer. 
We use both conditional generator and conditional discriminator to tackle multi-domain input and output. 
Our generator is inspired by the recent progress in arbitrary style transfer, and our discriminator is inspired by the recent progress in  generative adversarial networks. 
Our approach combines the best of both worlds. We propose a mask module that helps in both adversarial training and style transfer.
Moreover, we show that our trained discriminator can be used to select representative stylized image, which has been a long-standing problem.

Previous style transfer and GAN-based image translation methods only target on one domain, such as transferring the style of oil paint, or transforming from natural images to sketches. We systematically study the style transfer problem on a large-scale dataset of diverse artistic images. We can train one network to generate images in different styles, such as comic, graphite, oil paint, water color and vector art. Our approach generates more visually appealing results than previous style transfer methods, but there is still room to improve. For example, transferring image into 3D graphics with the arbitrary style transfer network is still challenging.



\printbibliography


\end{document}